\documentclass[conf]{new-aiaa}
\usepackage[utf8]{inputenc}

\usepackage{graphicx}
\usepackage{amsmath}
\usepackage{subcaption}
\usepackage{wrapfig}
\usepackage{bbm}
\usepackage{soul}

\usepackage{algorithm}
\usepackage{algorithmic}
\usepackage{multirow}

\usepackage{hyperref} 

\usepackage[version=4]{mhchem}
\usepackage{siunitx}
\usepackage{longtable,tabularx}
\usepackage{soul}
\usepackage{todonotes}
\usepackage{svg}

\title{Development and Testing for Perception Based Autonomous Landing of a Long-Range QuadPlane
}
\author{Ashik E Rasul\textsuperscript{*}}
\author{Humaira Tasnim\footnote{
Graduate Student, Mechanical Engineering Department, 1 William L Jones Dr, Cookeville, TN 38505.}}
\author{Ji Yu Kim\textsuperscript{\dag}\textsuperscript{\ddag}}
\author{Young Hyun Lim\footnote{Equal Contribution}\textsuperscript{\ddag}}
\author{Scott Schmitz\footnote{Undergraduate Student,  Mechanical Engineering Department, 1 William L Jones Dr, Cookeville, TN 38505.}}
\author{Bruce W. Jo\footnote{
Associate Professor, Mechanical Engineering Department, 1 William L Jones Dr, Cookeville, TN 38505.}}

\author{Hyung-Jin Yoon\footnote{
Assistant Professor, Mechanical Engineering Department, 1 William L Jones Dr, Cookeville, TN 38505.}}
\affil{Tennessee Technological University, Cookeville, TN 38505 USA}

\begin{document}

\maketitle

\begin{abstract} 
QuadPlanes combine the range efficiency of fixed-wing aircraft with the maneuverability of multi-rotor platforms, making them well-suited for long-range autonomous missions. In GPS-denied or cluttered urban environments, perception-based landing becomes essential for ensuring reliable operation. Unlike structured landing zones, real-world sites are often unstructured and highly variable, requiring strong generalization capabilities from the perception system. Deep neural networks (DNNs) offer a scalable solution for learning landing site features across diverse visual and environmental conditions. Although perception-driven landing has been successfully demonstrated in simulation, real-world deployment introduces significant challenges. Payload and volume constraints limit the integration of high-performance edge AI devices such as the NVIDIA Jetson Orin Nano Super, which are crucial for real-time object detection and control. Alongside perception-based control, accurate pose estimation during descent is necessary, especially in the absence of GPS, and relies on dependable visual-inertial odometry. Achieving this with the limited computational resources available on edge AI devices requires careful optimization of the entire deployment framework. The flight characteristics of large QuadPlanes further complicate the problem. These aircraft exhibit high inertia, reduced thrust vectoring capability, and slower response times, all of which contribute to the complexity of achieving precise and stable landing maneuvers.

This work presents the design, integration, and field testing of a lightweight QuadPlane framework capable of efficient vision-based autonomous landing and visual-inertial odometry, specifically developed for long-range QuadPlane operations such as aerial monitoring. It describes the hardware platform, sensor configuration, and embedded computing architecture designed to meet demanding real-time and physical constraints. This system establishes the foundation for future deployment of autonomous landing technologies in dynamic, unstructured, and GPS-denied environments. 
\end{abstract}

\section{Introduction}

\begin{figure}
    \centering
    \includegraphics[width=0.95\linewidth]{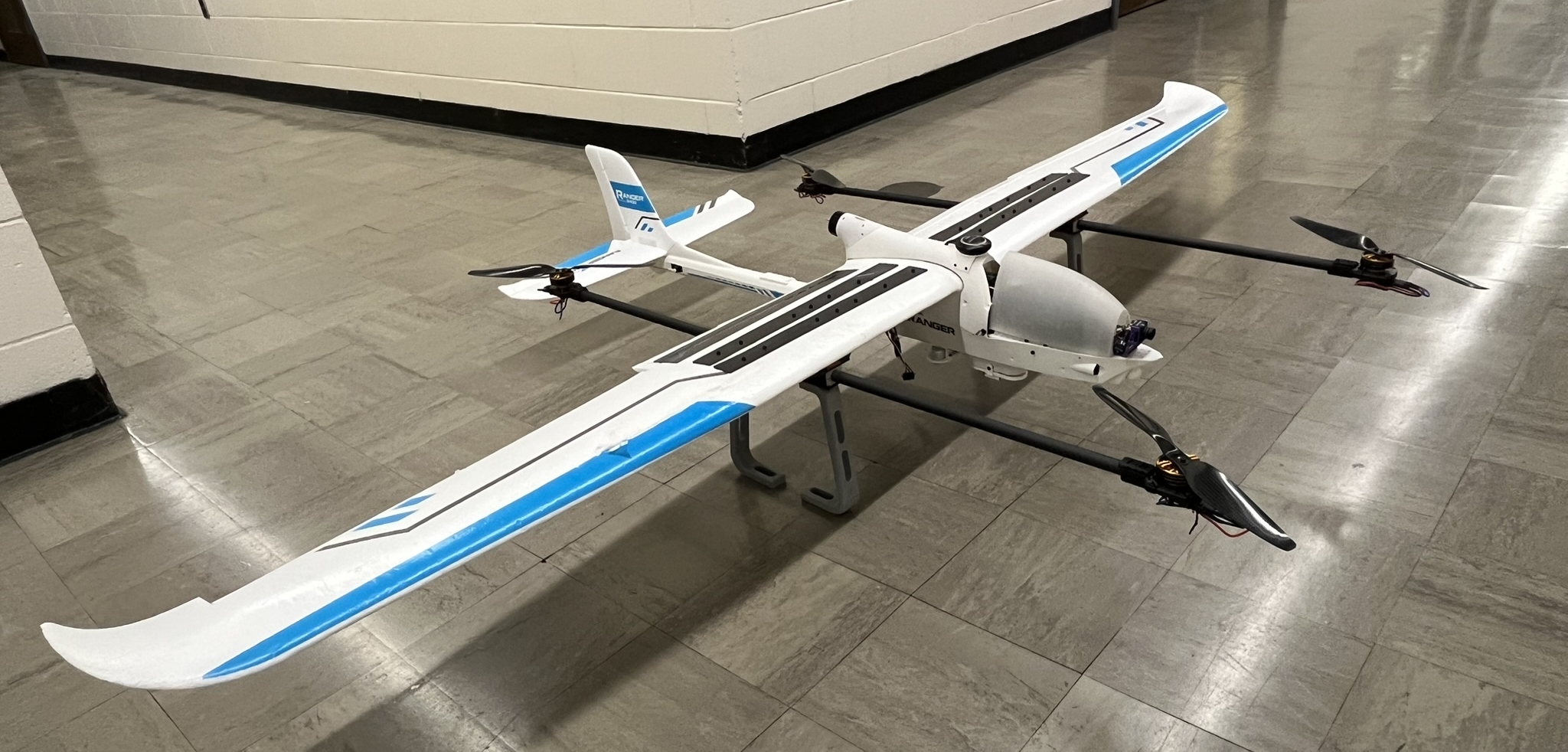}
    \caption{QuadPlane prototype featuring RGB, RGBD (Intel Realsense), FPV Cameras and Jetson Orin Nano}
    \label{fig:full)vtol}
\end{figure}
Deep Neural Networks (DNNs) demonstrate superior performance over traditional computer vision methods in a wide range of perception tasks~\cite{grigorescu2020survey}. General-purpose object detection frameworks such as YOLO~\cite{redmon2016,redmon2017yolo9000,ultralytics2020yolov5,ultralytics2023yolov8} exhibit strong capabilities in real-time object detection, especially when trained on domain-specific datasets~\cite{diwan2023object}. The performance of these detectors can be further improved through synthetic data augmentation and hyperparameter optimization techniques, including Bayesian optimization. Integration of these methods into autonomous aerial vehicle perception and control pipelines is successfully demonstrated in simulation environments~\cite{bansal2025airtaxisim,rasul2025bayesian,bansal2025verification}.  
Notably, these simulators leverage high-fidelity aerial vehicle dynamics engines such as GUAM~\cite{GUAM2024}. However, the implementation and validation of such systems in real-world scenarios remain an underexplored area of research.

Transferring perception-based landing systems from simulation to real-world QuadPlane platforms introduces three practical challenges. First, Training and refinement pipelines of DNN based perception models heavily depends on the realism of the simulation environment, such as CARLA~\cite{Dosovitskiy17}. Current implementation of such simulators~\cite{bansal2025airtaxisim} run on Unreal Engine (UE4)~\cite{Unreal}. To further enhance the realism, the simulator itself requires upgrades to more recent versions, such as Unreal Engine 5 (UE5). Second, the addition of sensors and edge AI devices~\cite{singh2023edge} capable of running real-time deep neural network inference must be achieved within strict payload and volume constraints. With the added system integration and onboard processing load, maintaining precise control and maneuverability of large-scale QuadPlanes becomes increasingly difficult due to their high inertia and limited thrust authority~\cite{mathur2021design}, which requires extensive optimization of the aerodynamic design \& validation with field testing. Third, the DNN model must be carefully optimized to fit within the limited computational budget of the onboard hardware~\cite{alqahtani2024benchmarking}, which must also share resources with visual-inertial odometry (VIO) required for accurate pose estimation during descent~\cite{nvidia2025isaacrosvslam}. 

In this work, we address the mentioned challenges through the design, integration, and field testing of a QuadPlane capable of performing computationally efficient vision-based landing and long-range QuadPlane operations such as aerial monitoring. The proposed system is built around a streamlined perception pipeline, which includes a YOLO-based object detector running in real-time on an NVIDIA Jetson Orin Nano Super edge AI device~\cite{nvidia2023orin_nano}. The model is compressed and optimized to meet the stringent latency and memory constraints of the hardware while preserving detection accuracy~\cite{ultralytics2024jetson}. To enable reliable tele-operation we integrate an first person view (FPV) camera. Moreover we integrate a  bottom facing RGB camera with adjustable focus to enable high speed sensing which is crucial for real time control and decision making. Furthermore, to enable the framework for reliable pose estimation through visual-inertial odometry~\cite{millane2024nvblox} during GPS-denied descent, we incorporate a compact depth camera, which shares computational resources with the object detector. The complete perception and VIO system is compactly integrated into the fuselage of a large (2.4 m wingspan) QuadPlane, adhering to strict payload and volume constraints as shown in figure \ref{fig:full)vtol}. Our design process leverages rapid prototyping techniques—specifically, CAD-based modeling and Fused Deposition Modeling (FDM) 3D printing for iterative development and mechanical integration, while ensuring aerodynamic balance, power efficiency, and robust sensor placement such that the added hardware does not compromise flight stability or control responsiveness. To the best of our knowledge, this is the first integrated hardware-software framework designed for onboard DNN-based autonomous landing of QuadPlanes in unstructured, GPS-denied environments, with planned full-scale field testing. Our key contributions are:
\begin{itemize}    
    \item We upgrade the iterative YOLO-based retraining pipeline for landing pad detection from CARLA UE4 to CARLA UE5, enabling improved realism and diversity in training and validation scenarios.
    
    \item We modify the design of a large-scale fixed-wing aircraft (2.4\,m wingspan with 4.5kg weight) for QuadPlane conversion, accommodating edge AI hardware and perception modules such as FPB camera, RGB Camera, RGBD camera and Arupilot based fligh control avionics. We leverage rapid prototyping tools such as CAD and FDM 3D printing \footnote{\url{https://github.com/ashikrasul/TTU_vtol}} to develop the reproducible framework followed by preliminary flight test. 
    
    \item We optimize the perception pipeline for real-time performance on the NVIDIA Jetson Orin Nano Super, including model compression and efficient resource sharing with the VIO module to meet tight computational and memory constraints.
\end{itemize}

\section{Literature Review}

Autonomous vertical landing of unmanned aerial vehicles (UAVs) is widely regarded as one of the most safety–critical phases of flight due to the strong coupling between perception, state estimation, and control under ground effect, aerodynamic disturbances, and limited thrust margins. Unlike fixed-wing runway landings, vertical take-off and landing (VTOL) requires precise localization of the touchdown zone directly beneath the aircraft, real-time obstacle awareness, and robust descent control at low altitude. Vision-based perception emerged as the dominant sensing modality for autonomous landing in GPS-denied and cluttered environments~\cite{gautam2014survey, xin2022visionlanding}. However, generalization of landing site understanding, onboard computation constraint, and robustness in various lighting and texture variations remain open challenges.

Early VTOL landing systems relied on cooperative visual markers placed on the landing surface. One of the first vision-based control frameworks for autonomous helicopter landing was demonstrated using image-based tracking and Kalman filtering \cite{saripalli2003vision}. Later, aggressive quadrotor landing was achieved  on a moving platform using onboard deep visual detection and nonlinear control \cite{falanga2017landing}. These works establish the feasibility of marker-based or helipad-based VTOL landing under motion and wind disturbances.

Markerless VTOL landing has gained increasing attention in recent years.  A vision-based optimal guidance law was developed for VTOL UAVs landing on moving platforms without relying on specially engineered landing markers \cite{jang2024vtol}. Another work proposed a deep-learning-based visual servoing framework for ship-deck VTOL recovery under significant deck motion and wave-induced disturbances \cite{cho2022autonomous}. RGB-D sensing further improves landing safety by providing direct geometric information near the ground. Real-time depth-variance-based VTOL landing on embedded GPUs is demonstrated \cite{paul2025autonomous}. These works demonstrate that deep perception along with depth sensing can replace handcrafted landing cues for VTOL platforms, but their evaluation remains limited to small multirotor systems.


Hybrid VTOL platforms aim to combine the hover capability of multirotors with the endurance of fixed-wing UAVs. The MiniHawk-VTOL platform demonstrates rapid prototyping of a tiltrotor UAV with validated hover-to-cruise transitions \cite{carlson2021minihawk}. While MiniHawk focuses on flight dynamics and control, it does not integrate onboard deep perception or autonomous landing logic. Modular UAV platforms, such as the sensor-centric architecture for environmental monitoring \cite{gu2018modularuav}, demonstrate how multi-sensor payloads and embedded processing can be integrated on ArduPilot-based UAVs. However, these platforms do not address real-time perception-based VTOL landing. In contrast to multirotors, QuadPlanes exhibit higher inertia, reduced thrust authority in hover mode, and stricter fuselage volume constraints for avionics and sensors. These characteristics significantly magnify the impact of perception latency and pose-estimation error during vertical descent. As a result, perception-based VTOL landing on large QuadPlanes remains largely underexplored in the literature.

High-fidelity simulation environments play a central role in developing VTOL landing perception. CARLA \cite{dosovitskiy2017carla} and AirSim \cite{airsim2017} have been widely used for synthetic data generation and control validation. However, the simulation-to-real gap remains a major challenge for VTOL perception systems. Recent work focused on an iterative retraining pipeline using CARLA-UE4 to improve YOLO-based landing detection under adverse weather and lighting \cite{rasul2025bayesian}. Nevertheless, Unreal Engine 4 limits photorealism and shadow fidelity. Upgrading the VTOL simulation pipeline to Unreal Engine 5 (UE5) introduces photorealistic lighting, improved global illumination, and richer scene geometry, which significantly narrows the domain gap between simulation and real-world VTOL operations. This enhancement is particularly important for helipad detection under complex illumination and background clutter.

Practical VTOL landing requires fully onboard perception and state estimation. To run YOLO detectors on NVIDIA Jetson platforms, TensorRT acceleration, FP16 inference, and model pruning are essential for real-time UAV perception. Together with depth-based visual inertial odometry (VIO), it is further required to run the inference pipeline with sub-20~ms latency on embedded GPUs. 

Accurate pose estimation during VTOL descent is equally critical. NVIDIA Isaac ROS Visual SLAM integrates RGB-D and IMU sensing to provide real-time VIO on embedded platforms \cite{nvidia2025isaacrosvslam}. However, few prior VTOL landing systems run both deep neural inference and visual–inertial odometry concurrently on a single low-power onboard processor.

From the VTOL-specific literature, three dominant gaps emerge: (i) existing VTOL landing simulation frameworks rely predominantly on UE4-level realism. (ii) most perception-based VTOL landing systems are demonstrated on small multirotors rather than long-range QuadPlanes; (ii) DNN-based landing pad detection and VIO are rarely co-optimized on a single embedded device. This work addresses these limitations by introducing a UE5-based iterative VTOL perception training pipeline, a structurally reinforced long-range QuadPlane platform, and a tightly optimized RGB + RGB-D + VIO perception stack deployed on an NVIDIA Jetson Orin Nano Super for real-time perception-based autonomous VTOL landing in GPS-denied environments.

\section{Proposed Framework}
Our proposed framework has three components: 

\subsection{Enhancing Iterative DNN Retraining Framework with CARLA UE5}

\begin{figure}[htbp]
    \centering
    \begin{subfigure}[b]{0.30\linewidth}
        \centering
        \includegraphics[width=\linewidth]{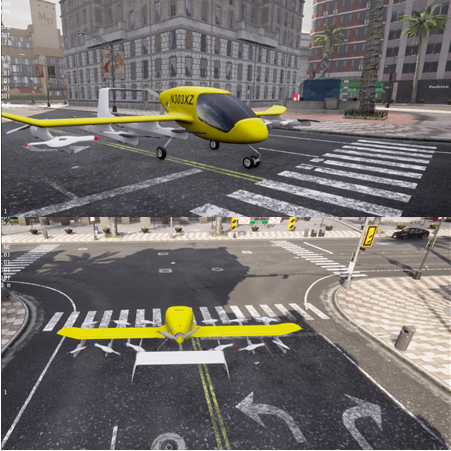}
        \caption{Photo-realistic simulator with Carla on UE4}
        \label{fig:vtol-avionics}
    \end{subfigure}
    \hfill
    \begin{subfigure}[b]{0.30\linewidth}
        \centering
        \includegraphics[width=\linewidth]{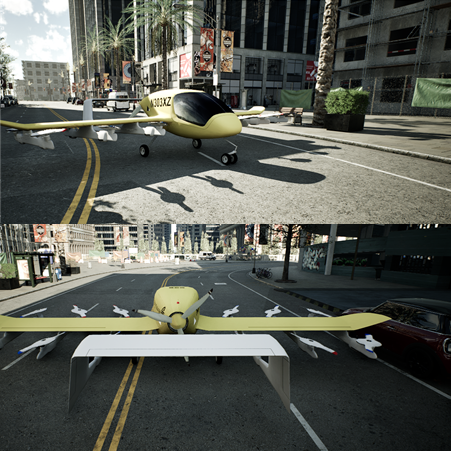}
        \caption{Photo-realistic simulator with Carla on UE5}
        \label{fig:vtol-avionics}
    \end{subfigure}
    \hfill
    \begin{subfigure}[b]{0.30\linewidth}
        \centering
        \includegraphics[width=\linewidth]{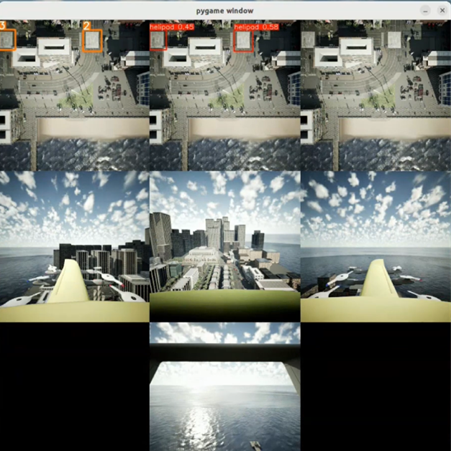}
        \caption{Perception-based landing in Carla UE5}
        \label{fig:vtol-prototype}
    \end{subfigure}
    \caption{Iterative improvement of perception-based model in Carla UE5}
    \label{ue5s}
\end{figure}

The performance of deep neural network (DNN) models for visual perception is strongly influenced by the realism of the training environment. Previous iterations of model refinement employed CARLA UE4 based AirTaxiSim simulator~\cite{bansal2025airtaxisim} in conjunction with Bayesian optimization~\cite{rasul2025bayesian}, yielding measurable improvements in detection performance in different weather and ligting conditions. With the release of CARLA UE5, which introduces advanced rendering features with enhanced photorealism (shown in Figure~\ref{ue5s}), we integrate our perception-based landing system within the upgraded simulator. This enables iterative fine-tuning of the pretrained model under more realistic and diverse conditions, improving its generalization to real-world scenarios. 

\subsection{QuadPlane Development for Real-World Deployment} In this section we describe different design and development steps of the QuadPlane framework. We first describe the design requirements considered for this development, then we describe the avionics and power distribution followed by restrengthening of the structure to enable the QuadPlane to carry additional load. 

\begin{figure}[htbp]
    \centering
    \begin{subfigure}[t]{0.45\linewidth}
        \centering
        \includegraphics[width=\linewidth]{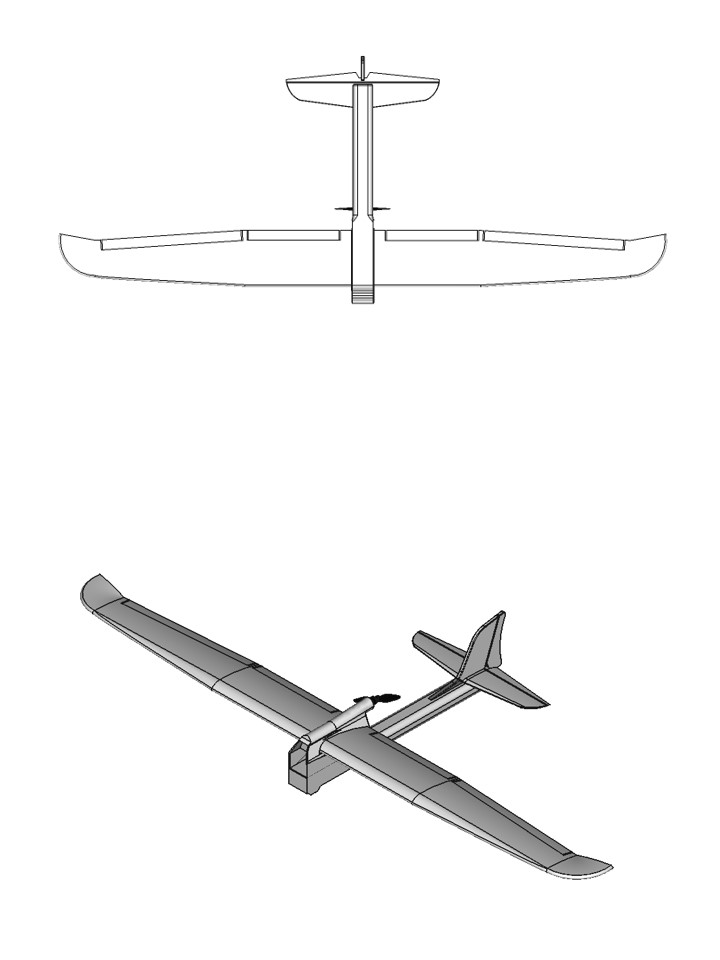}
        \caption{Unmodified fixed-wing frame}
        \label{voluntex}
    \end{subfigure}
    \hfill
    \begin{subfigure}[t]{0.45\linewidth}
        \centering
        \includegraphics[width=\linewidth]{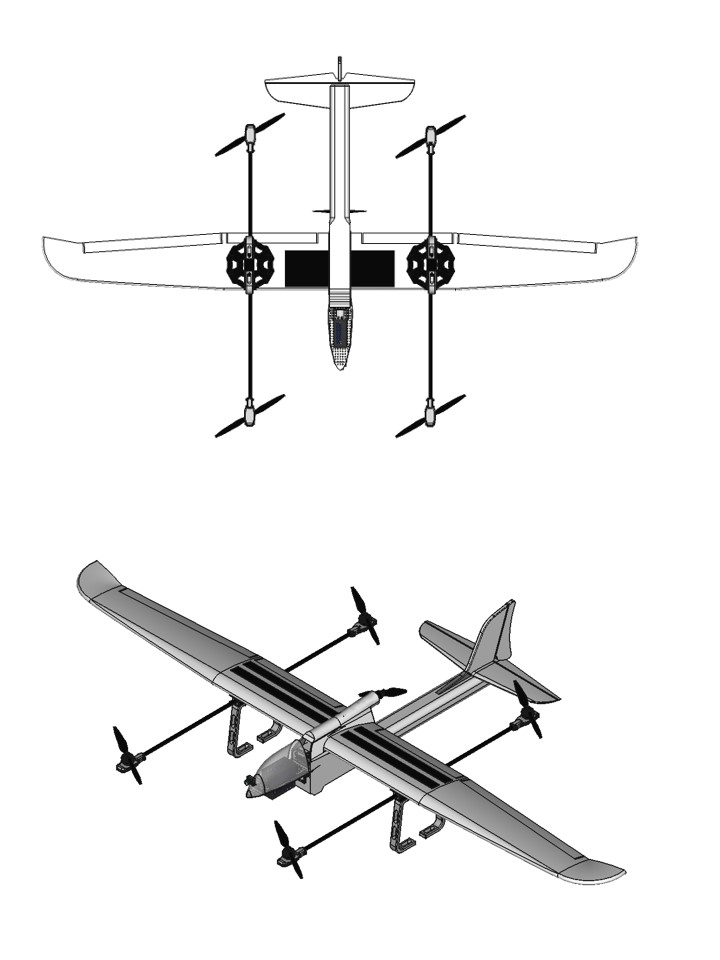}
        \caption{Modified QuadPlane}
        \label{frame_cad}
    \end{subfigure}
    \caption{QuadPlane frame design modification with integration of avionics, rotor arms, perception sensors and inference device with , structural restrengthening}

\end{figure}

\subsubsection{Design Requirements}

The following requirements were considered while designing the QuadPlane. 

\begin{enumerate}
    \item The QuadPlane main frame has to be long range glider type which enable it for high endurance aerial monitoring and exploration.
    \item The aircraft must have vertical take off and landing(VTOL) capability, that will enable it to manuver in cluttered urban environment. 
    \item The QuadPlane should have a load bearing capability of $\sim8$ kg, which will allow it to carry necessary perception, processing and actuation hardware.
    \item The fuselage should have enough space to accommodate the mentioned hardware. 
    \item The QuadPlane should accommodate at least 3 cameras along with the standard sensors: first person view (FPV), RGB and RGBD camera.
\end{enumerate}

\begin{figure}[htbp]
    \centering
    \begin{subfigure}[b]{0.45\linewidth}
        \centering
        \includegraphics[width=\linewidth]{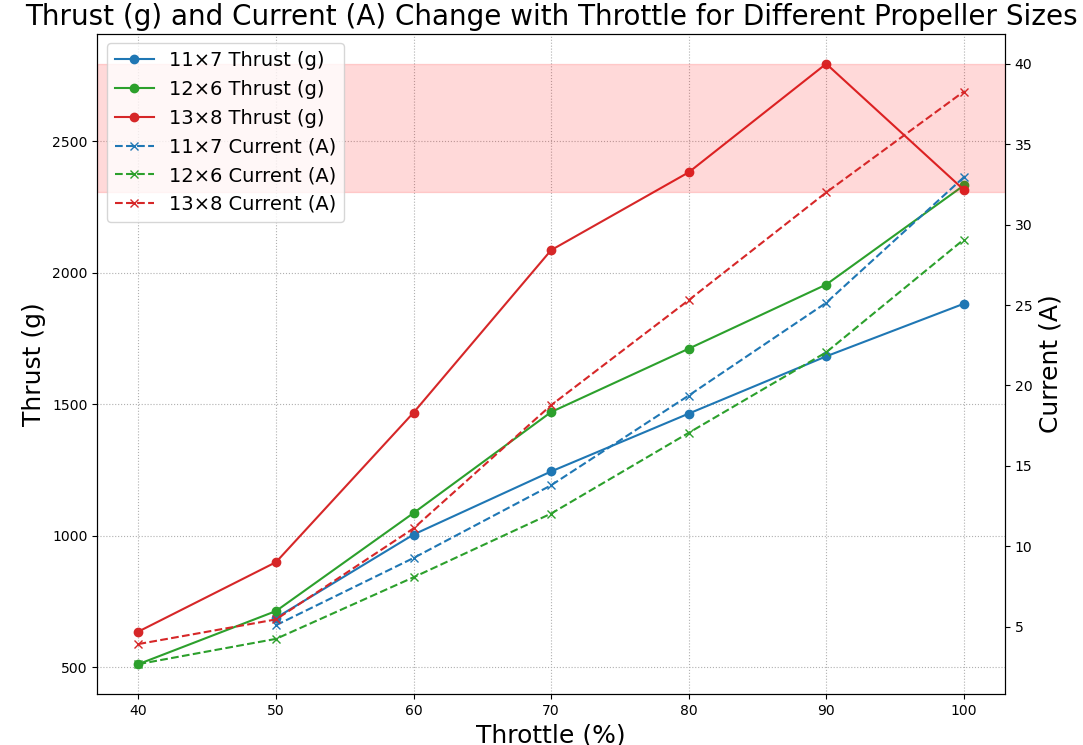}
        \caption{We select $12$ in propeller with $5.5$ in pitch due to safe operating in full throttle range providing 2kg thrust.}
        \label{thurst}
    \end{subfigure}
    \hfill
    \begin{subfigure}[b]{0.40\linewidth}
        \centering
        \includegraphics[width=\linewidth]{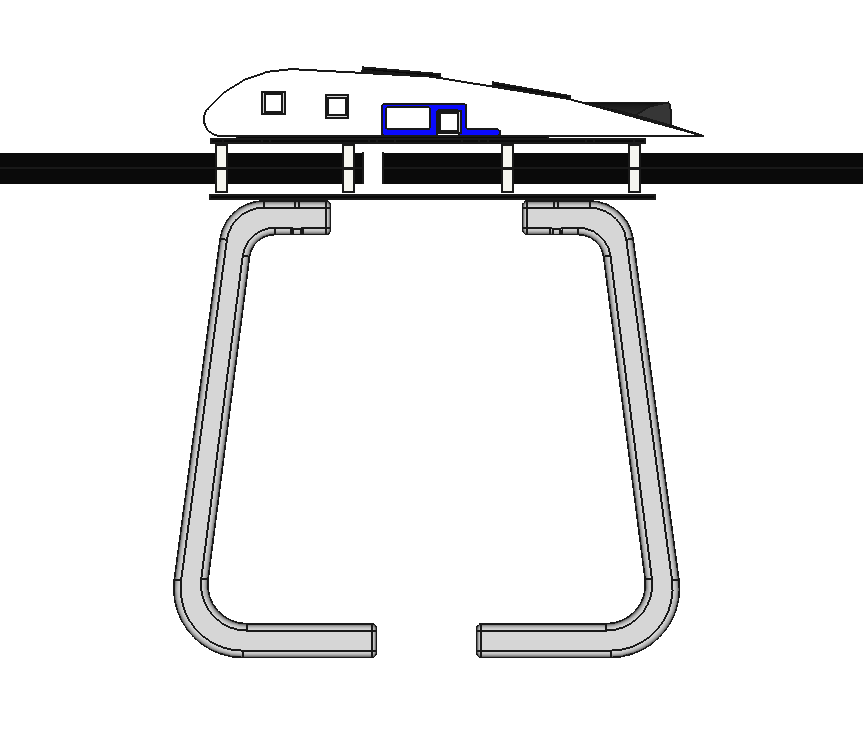}
        \caption{Rotor arms mounting with the fixed wing and integration of landing legs.}
        \label{props}
    \end{subfigure}
    \caption{Selection of motor and propellers to ensure proper thrust to weight ratio}
    \label{motor}
\end{figure}

We design our QuadPlane platform based on an already available fixed-wing frame \textit{Voluntex Ranger 2400}~\cite{volantex2025ranger2400} with a wingspan of $2.4$ meters as shown in figure \ref{voluntex}. We reconstruct the geometry in CAD (as illustrated in Figure~\ref{frame_cad}), enabling structural modifications and hardware integration. The total take-off weight of the platform is approximately $4.56$~kg. This includes the left and right wings (840~g each), fuselage (1450~g), battery (700~g), propellers (80~g), NVIDIA Jetson Orin Nano with protective casing (230~g), an Intel RealSense depth camera (120~g). To ensure a sufficient thrust-to-weight ratio for safe vertical takeoff, hover, and maneuverability, we select the QuadPlane motors \textit{Flash Hobby D4215}, each capable of delivering approximately $2$~kg of thrust, yielding a total thrust capacity of $8$~kg. We deploy $12$-inch propellers featuring a $5.5$-inch pitch which operates under $80\%$ of the maximum current of the ESC as shown in figure \ref{thurst}. Each motor is mounted on a $14$~mm diameter carbon fiber beam as shown in figure \ref{props}, with a length of $500$~mm from the aircraft's center of gravity. We develop two prototypes for this framework, Prototype~I and Prototype~II, respectively. 
Prototype~I is developed without the camera and Jetson module to achieve flight stability, and later Prototype~II integrates the perception sensors and the Jetson module.


\begin{figure}[htbp]
    \centering
    \begin{subfigure}[b]{0.45\linewidth}
        \centering
        \includegraphics[width=\linewidth]{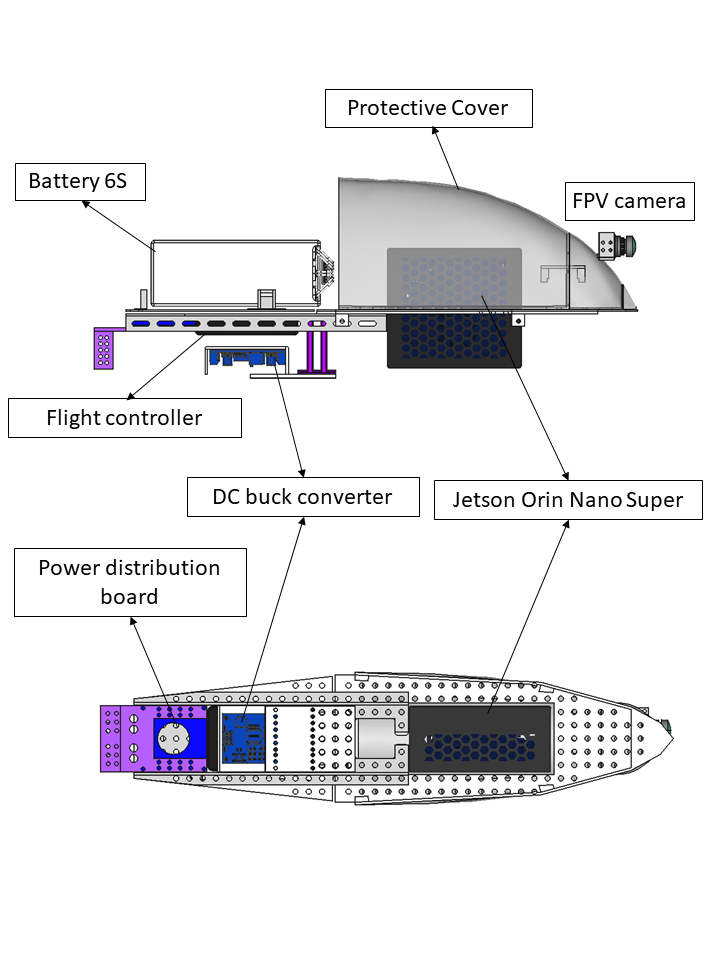}
        \caption{Distribution of avionics and Jetson components}
        \label{fig:vtol-avionics}
    \end{subfigure}
    \hfill
    \begin{subfigure}[b]{0.450\linewidth}
        \centering
        \includegraphics[width=\linewidth]{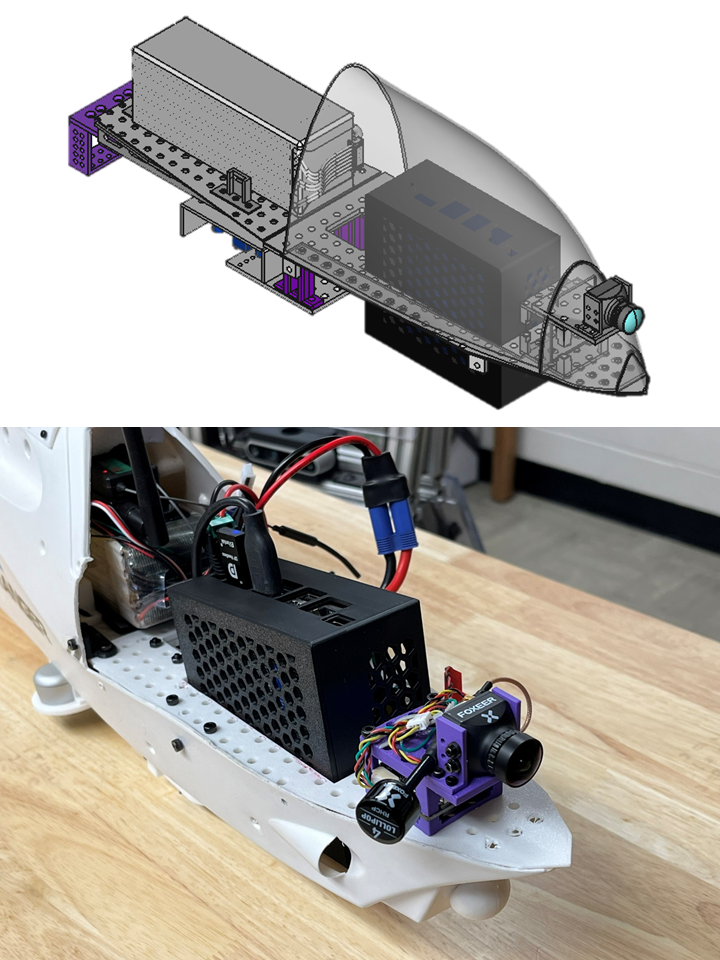}
        \caption{Design implementation in Prototype~II}
        \label{fig:vtol-prototype}
    \end{subfigure}

    \caption{Avionics and Edge-AI device integration}
\end{figure}

\subsubsection{Integration of Avionics and Edge-AI Device with Power Distribution}
We optimize the spatial layout of key components, such as the high-capacity LiPo battery, NVIDIA Jetson-based edge AI device, stereo/depth camera, flight controller, and power distribution board, through rapid prototyping using FDM 3D printing~\cite{cano2021fused}.  The flight control system is centered around the Pixhawk autopilot, integrated with Mission Planner as the ground station. A $6$S LiPo battery supplies power to the system via $40$~A ESCs. To support autonomous navigation and control in outdoor environments, we supplement the onboard inertial measurement unit (IMU) with an external GPS module and an airspeed sensor. 

In vision-based vertical landing systems for the QuadPlane, fast and robust perception is critical for accurate helipad detection during descent. Cloud-based inference pipelines introduce latency, reliance on wireless communication, and limited scalability, making them unsuitable for time-critical autonomous control tasks. To overcome these limitations, an onboard edge AI solution is required to ensure low-latency perception and autonomy. This work utilizes the NVIDIA Jetson Orin Nano Super~\cite{nvidia2023orin_nano}, a compact embedded platform that enables full on-device deep learning inference without external compute dependencies. The structural modification of the tight fixed wing fuselage (as shown in figure~\ref{fig:vtol-avionics}) allowed us to integrate the processing hardware required for running the DNN model inference (as shown in Figure~\ref{fig:vtol-prototype}) along with the avionics and power modules.  

\subsubsection{Restrengthening the Wings}
\begin{figure}[htbp]
    \centering
    \begin{subfigure}[b]{0.45\linewidth}
        \centering
        \includegraphics[width=\linewidth]{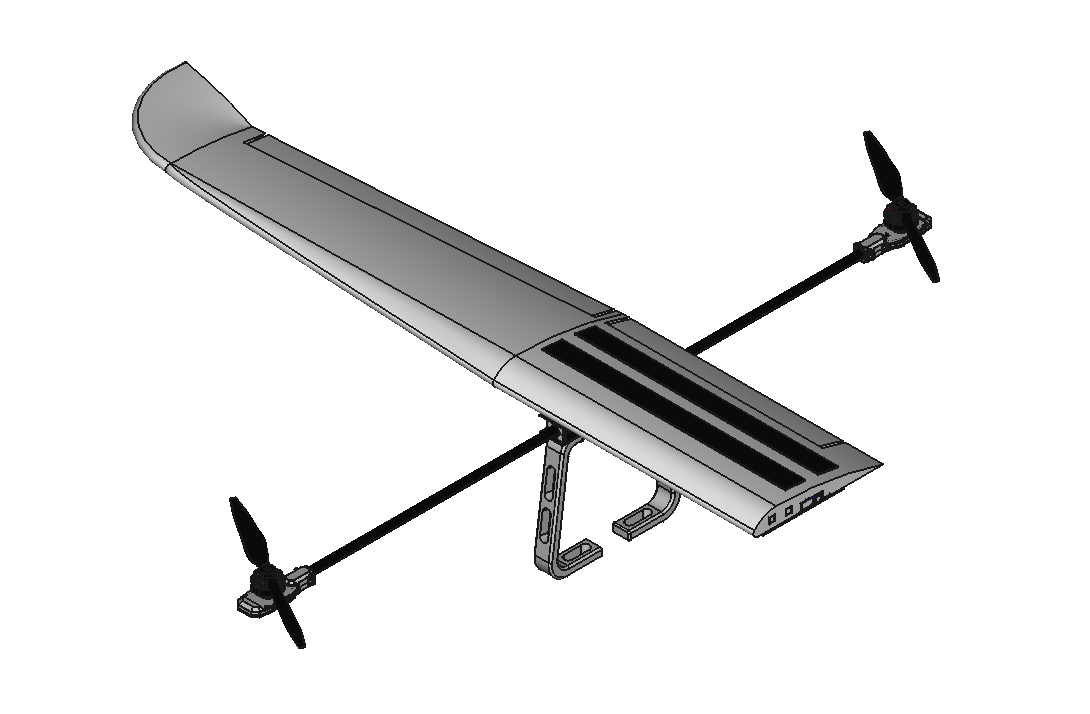}
        \caption{Additional carbon fiber plate and beam support}
        \label{wing_iso}
    \end{subfigure}
    \hfill
    \begin{subfigure}[b]{0.45\linewidth}
        \centering
        \includegraphics[width=\linewidth]{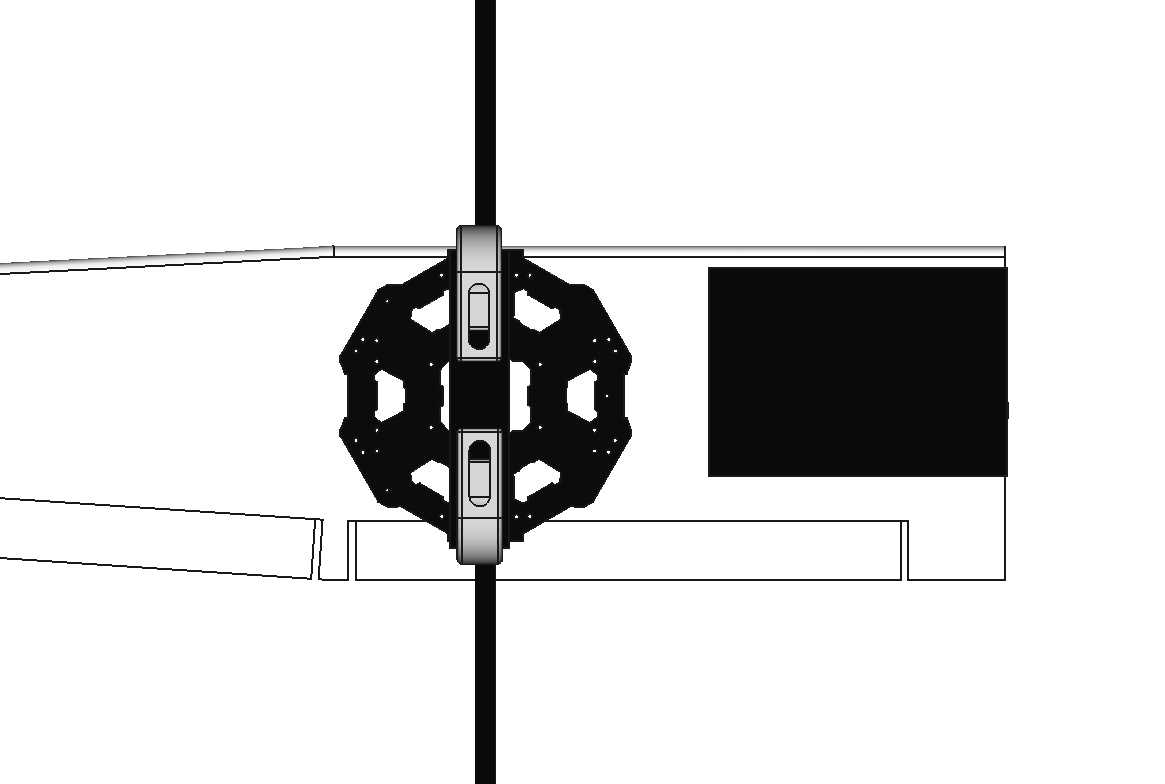}
        \caption{Mounting rotor arm and landing legs with carbon fiber beam support}
        \label{wing_back}
    \end{subfigure}
    \caption{Restrengthening of wings, rotor arm and landing leg mount}
    \label{fig:vtol-figures}
\end{figure}

We reinforce the QuadPlane wings with carbon fiber beams and plates, as shown in figure~\ref{wing_iso} to enhance structural rigidity and load-bearing capacity for the motor–arm–propeller assembly. A single 8 mm diameter carbon fiber beam, along with the existing two 8 mm beams, is embedded in each wing for this purpose. To secure the rotor arms to the wings, two additional 1.5 mm thick carbon fiber reinforcement plates are installed on the upper surface of each wing, ensuring sufficient stiffness during vertical takeoff and landing (VTOL) operations. Furthermore,  we embed 3D-printed inserts within the styrofoam wing core to provide mechanical support for the fasteners as they pass through the foam structure. Finally, we mount the motor-arm assembly with the wings with \textit{TAROT FY 690S} base plate~\cite{tarot_fy690s} as shown in figure~\ref{wing_back}. We maintain proper alignment of the aircraft center of gravity while accommodating the added structural reinforcements.

\subsection{Integration and Optimization of Perception Pipeline:}

\begin{figure}[H]
    \centering
    
    \begin{subfigure}[b]{0.30\linewidth}
        \centering
        \includegraphics[width=\linewidth]{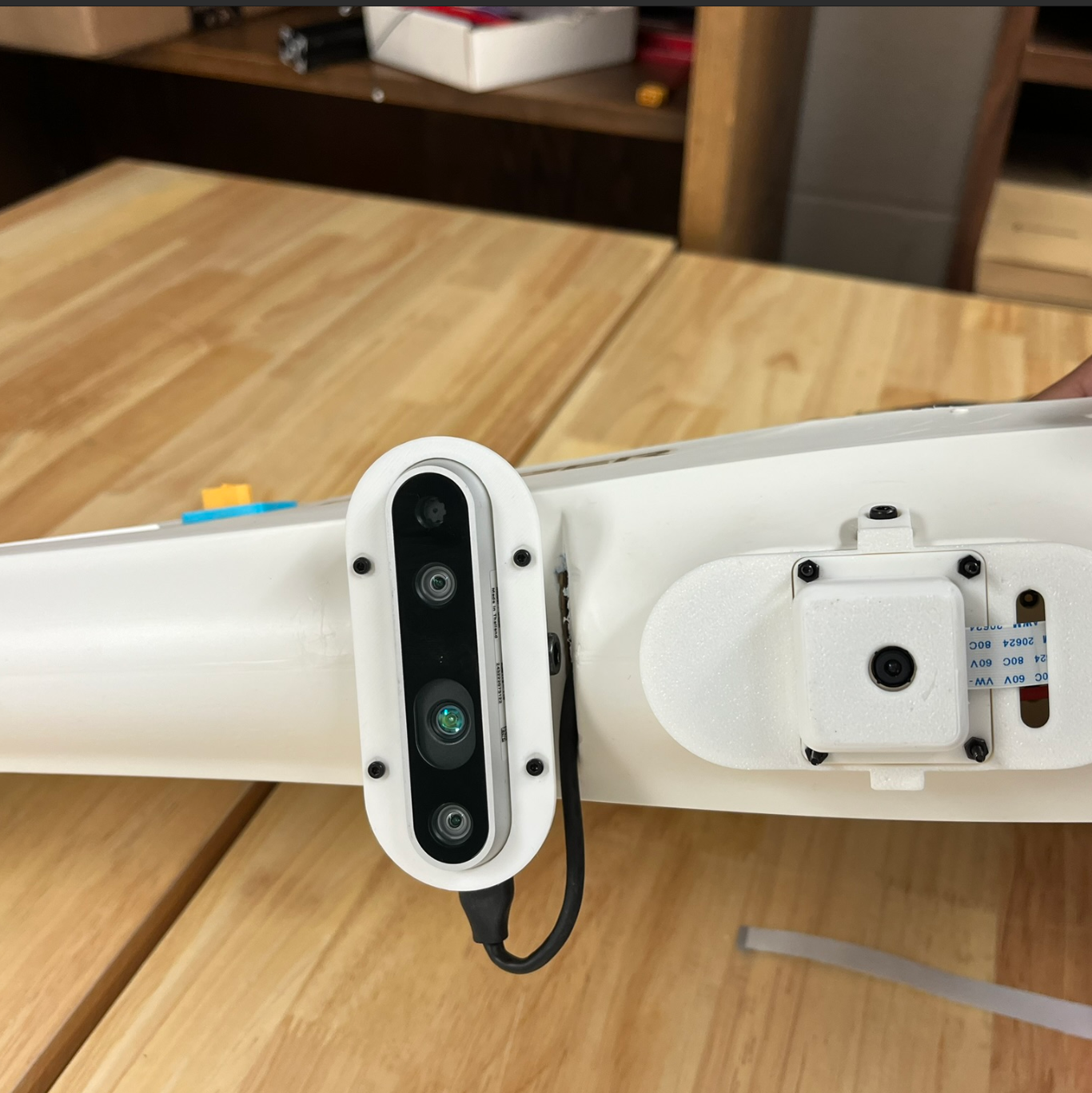}
        \caption{Integration of RGB and depth camera with QuadPlane}
        \label{fig:depth}
    \end{subfigure}
    \hfill
    \begin{subfigure}[b]{0.30\linewidth}
        \centering
        \includegraphics[width=\linewidth]{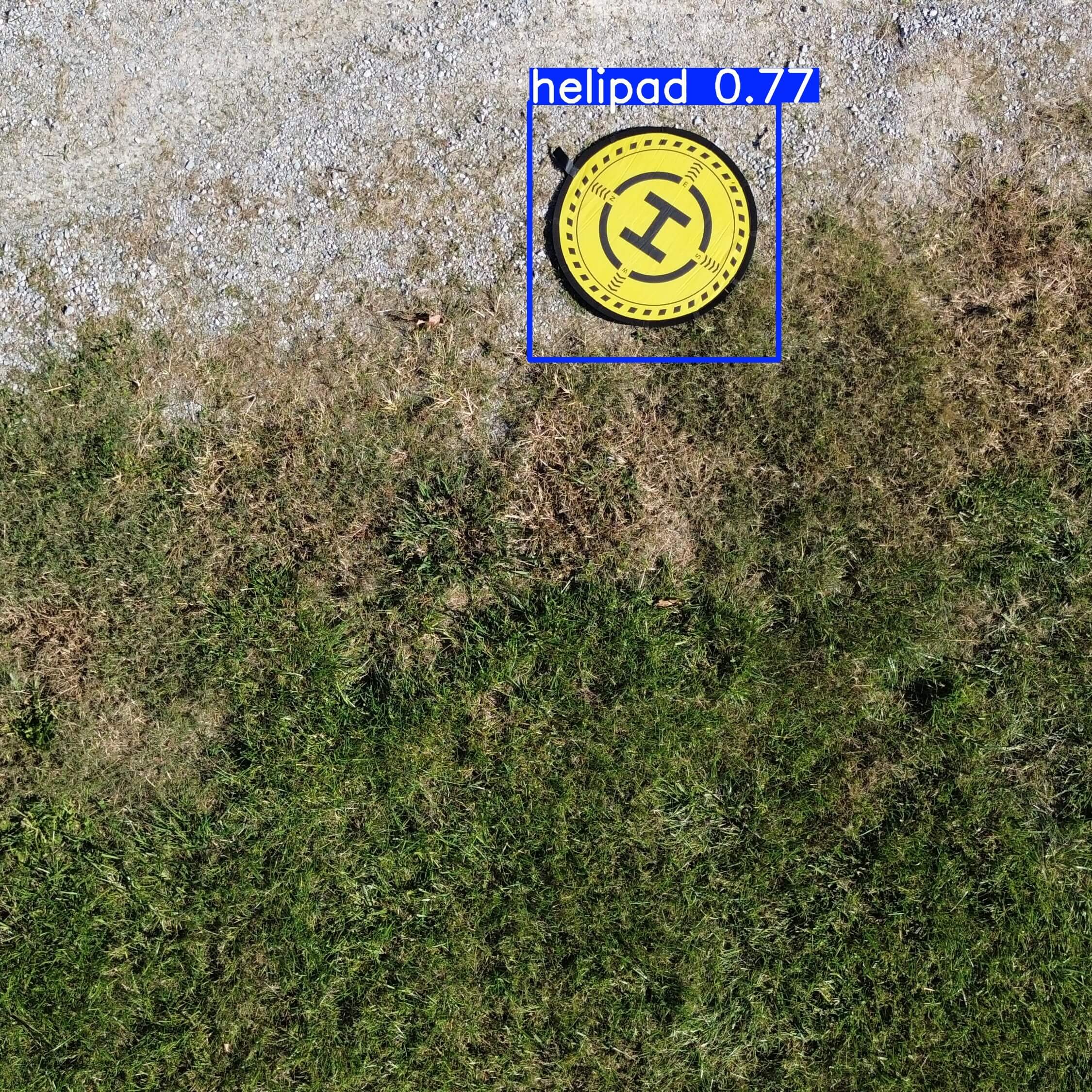}
        \caption{Helipad detection with TensorRT on Jetson Orin Nano}
        \label{fig:yolo}
    \end{subfigure}
    \hfill
    \begin{subfigure}[b]{0.30\linewidth}
        \centering
        \includegraphics[width=\linewidth]{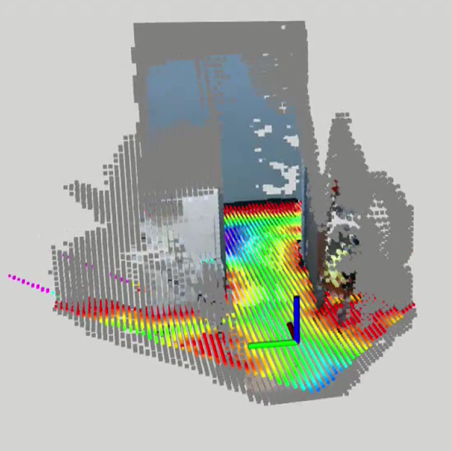}
        \caption{3D mapping with visual odometry using NVblox}
        \label{fig:nvblox}
    \end{subfigure}
    \caption{Demonstration of optimized perception framework with DNN inference and Visual Inertial Odometry (VIO)}
    \label{}
\end{figure}

We integrate an \textit{Arducam IMX519}~\cite{arducam_imx519} RGB camera for high-speed image acquisition and low-latency communication to enable real-time control and decision-making. In addition to the RGB camera, an RGB-D sensor, specifically the Intel RealSense D435i~\cite{realsense_d435i}, is incorporated. Both cameras are mounted in a downward-facing configuration beneath the aircraft fuselage, as shown in Fig.~\ref{fig:depth}.

For real-time deployment of the perception model the trained PyTorch model (\texttt{.pt}) is first exported to the ONNX format using the Ultralytics exporter. The resulting ONNX model is then converted into a TensorRT engine with FP16 precision to minimize inference latency and memory footprint. The obtained TensorRT engine is evaluated on a dataset of 120 unseen helipad images collected under diverse outdoor conditions, as shown in Figure~\ref{fig:yolo}. We could achieve an average processing time of 8.5~ms per image for preprocessing, followed by 19.1~ms for GPU-based inference using FP16 precision, and 5.1~ms for postprocessing, including non-maximum suppression and decoding. This results in a total per-frame inference time of approximately 32.7~ms, enabling real-time operation at over 30 frames per second (FPS). 

Reliable pose estimation is critical for autonomous landing in GPS-denied environments. To achieve this, we implement a visual-inertial odometry (VIO) pipeline optimized for the Jetson Orin Nano Super. The compact depth camera provides synchronized RGB-D input as shown in while an onboard IMU delivers acceleration and angular velocity data. These sensor streams are fused using NVIDIA’s \texttt{Isaac ROS Visual SLAM}~\cite{nvidia2025isaacrosvslam}, producing real-time 6-DoF pose estimates of the vehicle in the world frame, independent of GNSS signals. To enhance spatial awareness, the VIO output is integrated with \texttt{nvblox}~\cite{millane2024nvblox} to incrementally construct a voxel-based map of the environment as shown in Figure~\ref{fig:nvblox}. The entire VIO and estimation pipeline operates onboard, ensuring real-time responsiveness and autonomy without dependence on external infrastructure.

\section{Flight Test and Results}
\begin{figure}[H]
    \centering

    \begin{subfigure}[t]{0.40\linewidth}
        \centering
        \includegraphics[width=\linewidth]{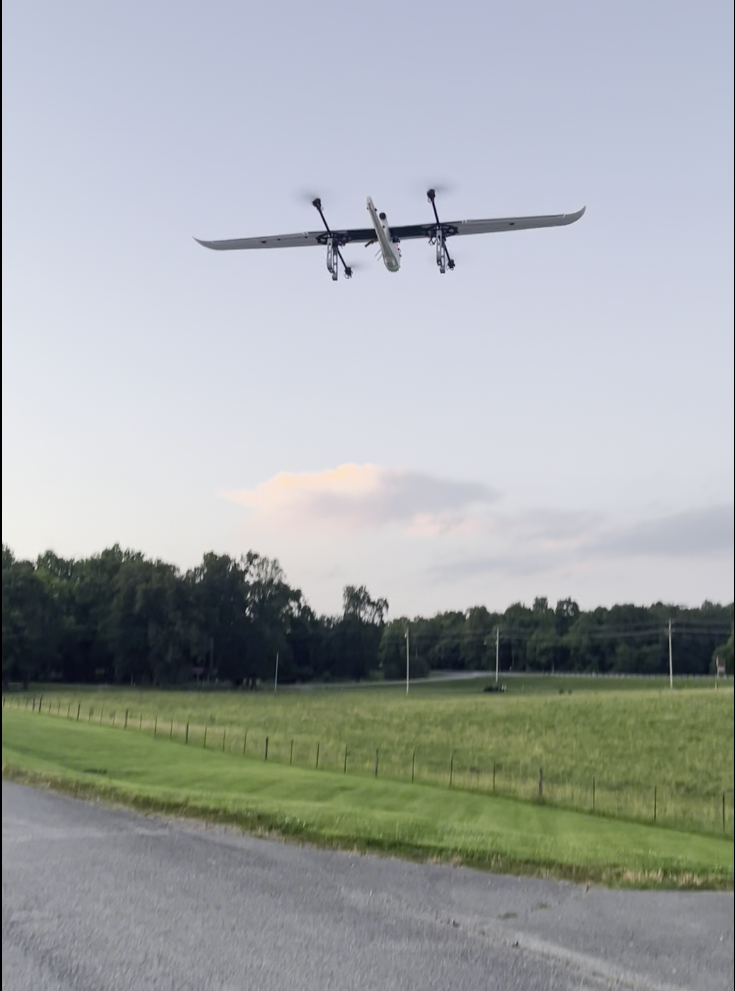}
        \caption{Flight Test with Loiter Mode}
        \label{fig:flight_test}
    \end{subfigure}
    \hfill
    \begin{subfigure}[t]{0.40\linewidth}
        \centering
        \includegraphics[width=\linewidth]{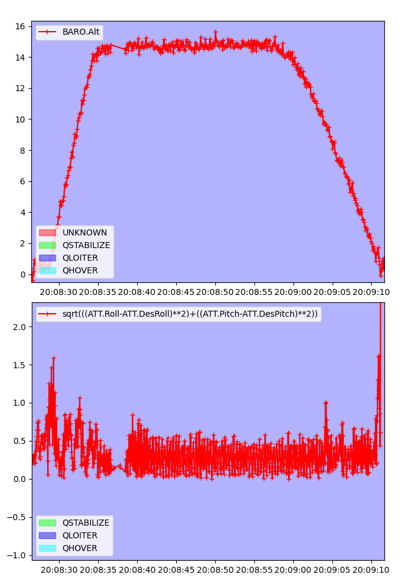}
        \caption{(Top) Altitude hold performance showing showing $30s$ position hold, (Bottom) angular attitude error.  }
        \label{flight perform}
    \end{subfigure}

\end{figure}

Initial flight tests have been conducted \footnote{\url{https://www.youtube.com/watch?v=OYJIW1vKeb8}} using the QuadPlane platform without the edge AI device and the perception modules as shown in figure \ref{fig:flight_test}.  We have acquired stable loiter performance with prototype-1. The aircraft took off and reached an altitude of $15m$. It held the same altitude for $30s$ followed by landing. We observed the angular error in roll and pitch angle against the desired values as square root and found stable performance within $1.5$ degree as shown in figure \ref{flight perform}.

\section{Conclusion and Future Work}
In this work, we developed a QuadPlane platform capable of iterative retraining for perception-based autonomous landing within a state-of-the-art simulation environment. We performed comprehensive hardware design, modification, and system integration to transform a fixed-wing aircraft into a platform supporting perception-based autonomous landing with tightly coupled control, perception, and deep neural network (DNN) processing modules. We further demonstrated an optimized onboard inference pipeline for real-time decision-making using an edge AI device under resource-constrained conditions. Finally, the baseline flight stability of the QuadPlane was validated through loiter-mode flight tests using a simplified prototype configuration.

This work is ongoing, and we plan to revalidate the full flight envelope through further field tests along with numerical analysis of the aerodynamic design~\cite{islam2025design} using the already developed Prototype~II ,which incorporates the NVIDIA Jetson Orin Nano Super and the complete perception module suite. These tests will ensure that the added payload does not compromise control responsiveness or flight stability. While the depth camera has been functionally verified in ground experiments, its performance and data integrity must now be evaluated under actual flight conditions. Upon confirmation of stable operation with all components onboard, a full-system autonomous flight test will be conducted to validate real-time helipad detection, onboard pose estimation via visual–inertial odometry, and controlled descent in a GPS-denied environment.

\bibliography{mybib}

\end{document}